\documentclass{article} 
\usepackage{iclr2026_conference,times}

\usepackage{amsmath,amsfonts,bm}

\def\eqref#1{equation~\ref{#1}}

\def\1{\bm{1}}

\DeclareMathAlphabet{\mathsfit}{\encodingdefault}{\sfdefault}{m}{sl}
\SetMathAlphabet{\mathsfit}{bold}{\encodingdefault}{\sfdefault}{bx}{n}

\usepackage{hyperref}
\usepackage{url}
\usepackage{minitoc}
\usepackage{inconsolata}
\usepackage{url}
\usepackage[linesnumbered,ruled,vlined]{algorithm2e}
\usepackage{multirow}
\usepackage{booktabs}
\usepackage{caption}
\usepackage{subcaption}
\usepackage{xspace}
\usepackage{amsmath}
\usepackage{amssymb}
\usepackage{bbm}
\usepackage{graphicx}
\usepackage{tabularx}
\usepackage{bbding}
\usepackage{pifont}
\usepackage{diagbox}
\usepackage{makecell}
\usepackage{colortbl}
\usepackage{bbm}
\usepackage{color}
\usepackage{hhline}
\usepackage{soul}
\usepackage{float}
\usepackage[most]{tcolorbox}
\usepackage[framemethod=tikz]{mdframed}
\usepackage{lipsum}
\usepackage{wrapfig}

\title{Low-rank Optimization Trajectories Modeling for LLM RLVR Acceleration}
\iclrfinalcopy

\author{\textbf{Zhipeng Chen\textsuperscript{{1}},~
        Tao Qian\textsuperscript{{2}},~
        Wayne Xin Zhao\textsuperscript{{1}}\thanks{\llap{}\:\:\:Corresponding authors.},~
        Ji-Rong Wen\textsuperscript{{1}}
    } \\
    \textsuperscript{1}Gaoling School of Artificial Intelligence, Renmin University of China.\\
    \textsuperscript{2}China University of Mining and Technology (Beijing). \\
    \texttt{zhipeng\_chen@ruc.edu.cn, batmanfly@gmail.com}
}

\newcommand{\paratitle}[1]{\vspace{1.5ex}\noindent\textbf{#1}}
\newcommand{\ie}{\textit{i.e.,}\xspace}

\newcommand{\eg}{\textit{e.g.,}\xspace}

\newcommand{\ignore}[1]{}

\definecolor{takeaway}{RGB}{230,230,250}
\definecolor{takeawayTitle}{RGB}{57, 89, 163}

\begin{document}

\maketitle

\begin{abstract}
Recently, scaling reinforcement learning with verifiable rewards (RLVR) for large language models (LLMs) has emerged as an effective training paradigm for significantly improving model capabilities, which requires guiding the model to perform extensive exploration and learning, leading to substantial computational overhead and becoming a key challenge.
To reduce the number of training steps, Prior work performs linear extrapolation of model parameters.
However, the dynamics of model parameter updates during RLVR training remain insufficiently understood.
To further investigate the evolution of LLMs during RLVR training, we conduct empirical experiments and find that the rank-1 subspace of the model does not evolve linearly, and its dominance over the original parameters is further amplified during LoRA training.
Based on the above insights, we propose the \textbf{N}onlinear \textbf{Ext}rapolation of low-rank trajectories (\textbf{NExt}), a novel framework that models and extrapolates low-rank parameter trajectories in a nonlinear manner.
Concretely, we first train the model using LoRA and extract the rank-1 subspace of parameter differences at multiple training steps, which is then used for the subsequent nonlinear extrapolation.
Afterward, we utilized the extracted rank-1 subspace to train a predictor, which can model the trajectory of parameter updates during RLVR, and then perform the predict-extend process to extrapolate model parameters, achieving the acceleration of RLVR.
To further study and understand NExt, we conduct comprehensive experiments that demonstrate the effectiveness and robustness of the method.
Our method reduces computational overhead by approximately 37.5\% while remaining compatible with a wide range of RLVR algorithms and tasks.
We release our code in \url{https://github.com/RUCAIBox/NExt}.
\end{abstract}

\section{Introduction}

Reinforcement learning with verifiable rewards (RLVR) can enhance the reasoning ability of large language models (LLMs), enabling them to engage in more thorough and structured thinking during the reasoning process~\cite{kimi-k1.5,deepseek-r1,openai-o1}.
During RLVR training, the model is required to conduct extensive exploration and learn from the experiences obtained through this exploration~\cite{still-3}.
However, this reliance on large-scale exploration fundamentally limits the scalability of RLVR, turning it into a computational bottleneck that restricts further capability gains of LLMs.
As model sizes and reasoning complexity continue to grow, this cost is not merely inconvenient, which becomes prohibitively expensive and increasingly unsustainable.

To accelerate the RLVR training process, existing work has primarily focused on optimizing the training procedure, including selecting data with higher learning potential~\cite{Zhu-arxiv-2025-The,Tang-arxiv-2025-Rethinking}, designing more effective exploration strategies~\cite{Huang-arxiv-2025-Low,Yang-arxiv-2025-Depth}, and developing more appropriate reward functions~\cite{passk_training,still-4}.
These approaches treat LLM optimization as a black-box process, focusing on improving sampling efficiency or reward design, while leaving the inherently iterative nature of RLVR untouched.
As a result, they can only provide marginal improvements, but fail to address the core inefficiency rooted in repeated exploration and update cycles.
This raises a more fundamental question: \emph{Is it necessary to follow the entire RL trajectory step by step, or can we directly predict its outcome?}

\begin{figure}[t]
    \centering
    \includegraphics[width=0.9\linewidth]{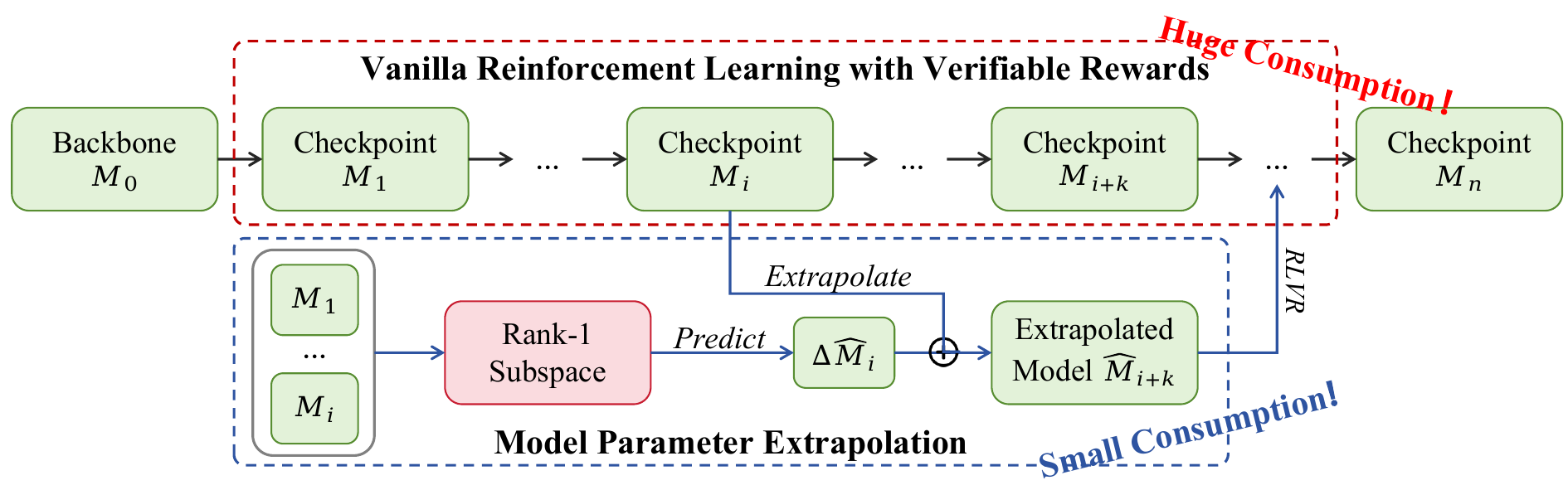}
    \caption{Comparison between vanilla RLVR and model parameter extrapolation. The vanilla RLVR consumes huge computational resources, while the extrapolation method can skip the intermediate training steps and predict the model parameters in the future checkpoint.}
    \label{fig:intro}
\end{figure}

However, directly predicting the full set of model parameters after RLVR is prohibitively difficult, due to the extreme dimensionality and highly non-linear dynamics of LLM optimization, making naive prediction approaches fundamentally infeasible.
To mitigate this difficulty, recent work (\eg AlphaRL~\cite{alpha_rl}, RL-Extra~\cite{rl_extra}) proposes to approximate the parameter updates using a rank-1 subspace and performs linear extrapolation within this subspace to predict the post-RLVR model, thereby reducing the required training steps.
Despite their empirical success, these approaches rely on a strong yet underexplored assumption: the dominant direction captured by the rank-1 subspace is sufficient to characterize the entire RLVR-induced parameter transformation. 
The validity of this assumption, as well as the underlying dynamics of parameter updates during RLVR, remains insufficiently explored.

Motivated by this, we conduct an empirical investigation into the evolution of LLM parameters during RLVR training, with a particular focus on the behavior of the rank-1 subspace. Our analysis reveals two important observations. First, the rank-1 subspace exhibits increasingly dominant influence over parameter updates during training, especially under LoRA-based fine-tuning. Second, the evolution of the rank-1 subspace does not strictly follow a linear pattern, indicating that linear extrapolation may be insufficient to accurately capture the underlying dynamics. These findings provide new insights into the structure of RLVR optimization and suggest the necessity of more expressive modeling approaches.

Based on the above observations, we propose \textbf{N}onlinear \textbf{Ext}rapolation of low-rank Trajectories (\textbf{NExt}), a novel framework that models and extrapolates low-rank parameter optimization trajectories in a nonlinear manner, enabling direct prediction of future model states.
Specifically, we first perform RLVR training with LoRA and extract the rank-1 subspace of parameter differences at multiple training steps. 
We then construct a predictor to model the optimization trajectory of these low-rank representations, and employ a predict-extend paradigm to extrapolate model parameters toward future states. 
In this way, NExt reduces the need for exhaustive intermediate training steps and improves overall training efficiency.
To evaluate the effectiveness of the proposed method, we conduct comprehensive experiments across LLMs of different scales. 
The results demonstrate that NExt can significantly reduce computational overhead (by approximately $37.5$\%) while maintaining or even improving model performance. 
Furthermore, the method shows strong robustness with respect to different hyperparameters, downstream tasks, and RLVR algorithms.
Our main contributions are summarized as follows:

\begin{itemize}
    \item We provide the analysis of LLM parameter optimization trajectories during RLVR, uncovering that LoRA fine-tuning can better elicit the LLM rank-1 subspace, and not all rank-1 subspaces satisfy the linear transformation. These insights challenge the linearity assumption in prior work and provide new perspectives for designing acceleration methods for RLVR. (Section~\ref{sec:empirical_study})
    \item Based on the insights from our experiments, we propose nonlinear extrapolation of low-rank trajectories (\textbf{NExt}), which first models the low-rank parameter optimization trajectories of the LLM RLVR process and then leverages it to extrapolate the model parameters, thereby reducing the consumption of computational resources. (Section~\ref{sec:approach})
    \item We conduct experiments on four models of varying scales, and observe that the LLMs trained through NExt can achieve better performance than the model trained through vanilla RLVR, with a reduction in time cost by 37.5\%. Besides, the detailed analysis demonstrates that NExt is insensitive to the choice of hyperparameters, downstream tasks, and the backbone RLVR algorithm, highlighting its effectiveness, robustness, and generality. (Section~\ref{sec:experiment})
\end{itemize}

\section{Related Work}

\paratitle{Reinforcement Learning for LLMs.}
Reinforcement learning has become a critical training stage in LLM post-training~\cite{llmsurvey}, to align LLMs to human preference and enhance the capacities of LLMs.
For alignment, LLMs are guided to generate several responses based on the given prompt, and then these responses are evaluated by human feedback~\cite{Christiano-NeurIPS-2017-Deep,instructgpt}.
The collected feedback can be leveraged to optimize LLM parameters through RL algorithms, \eg PPO~\cite{ppo,secret_of_rlhf_1}
To reduce training costs, previous works train a reward model using human-annotated preference data, which provides reward signals during RL training~\cite{secret_of_rlhf_2}.
Given the challenges in reward modeling, existing studies directly optimize the model using positive and negative examples~\cite{dpo,simpo,allo}.
For enhancing LLM capacities, reinforcement learning with verifiable rewards (RLVR) has been widely applied to enhance the reasoning capabilities of large language models~\cite{deepseek-r1,kimi-k1.5,openai-o1}.
RLVR improves the model’s performance on diverse tasks through outcome-level reward signals~\cite{still-3,logic_rl,simple_rl_zoo,enigmata}.
Moreover, Since outcome-level rewards cannot fully evaluate the quality of model-generated responses, prior work has proposed methods such as process-supervised reward models~\cite{verify_step_by_step,math_shepherd,rlmec} and reverse curriculum reinforcement learning~\cite{Xi-icml-2024-Training} to provide finer-grained reward signals for training large language models.
In this work, we focus on employing the RLVR process to enhance the capacities of LLMs without extensive resource consumption.

\paratitle{RLVR Acceleration.}
Due to the substantial training time and computational resources required by RLVR, which is often exacerbated by the iterative rollout and reward feedback loops inherent in reinforcement learning paradigms~\cite{Venkatkrishna-arxiv-2025-Aletheia,roll_flash}, how to accelerate its training process has become a critical issue that restricts the practical deployment and scalability of RLVR-based LLM post-training.
To accelerate the RLVR, researchers filter the training data to retain only samples that are beneficial for learning~\cite{tang2025towards,zhu2025data,Chimoto-acl-2024-critical}, or weight the training instances according to their importance, which helps prioritize valuable samples and reduce the impact of redundant or noisy data~\cite{still-4,forking_token}.
However, since the data selection process itself incurs additional overhead, such as the cost of evaluating sample utility or designing effective filtering criteria~\cite{Wang-arxiv-2026-OPUS,Liu-naacl-2025-Take}. 
Recent studies have proposed lightweight utility estimation frameworks that reduce the computational burden of sample selection while maintaining effectiveness~\cite{zhou2026efficient,yan2024efficient}.
Furthermore, reusing sampled trajectories by storing and reprocessing historical rollout data can avoid redundant environment interactions and reward calculations~\cite{zhang2025improving,ren2026recycling,rlvr_world}, while leveraging offline data collected from prior training processes or public datasets can reduce the need for extensive online rollout processes~\cite{luffy,a2d}.
In this work, we focus on accelerating the training process from the perspective of model parameters rather than algorithmic optimization. 
Thus, our approach is orthogonal to prior work discussed above, and these methods are compatible with ours in principle.

\paratitle{LLM Parameter Extrapolation.}
Model parameter extrapolation is one approach to improving model capability and training efficiency~\cite{model_merging_survey_1,model_merging_survey_2}.
This method essentially leverages the inherent patterns and correlations within model parameters, either across training stages, across model scales, or across different checkpoint states, to predict or derive target parameters~\cite{Fei-arxiv-2022-MetaEnsemble,Knyazev-neurips-2021-Parameter}.
To improve training efficiency, prior work has designed linear extrapolation methods that leverage the evolution patterns of model parameters during training to predict parameters in later stages, thereby reducing training time~\cite{rl_extra,alpha_rl,zheng2025model}.
Additionally, linear extrapolation of gradients from different data batches guides training toward more robust model parameters, further optimizing training efficiency~\cite{Asaad-arxiv-2025-Gradient,Lin-icml-2020-Extrapolation}
Besides, several studies utilize the information from the well-trained small model to predict the parameters of the trained larger model~\cite{proxythinker,proxy_tuning}.
To improve training effectiveness, prior work has merged parameters from intermediate checkpoints saved during training to obtain a model with enhanced capability~\cite{wsm,sanyal2023early}.
Moreover, to avoid disrupting the model’s internal knowledge, existing work performs extrapolation on only a subset of parameters, thereby improving capability while preventing model collapse~\cite{maec,yu2024language,seallms3}.
In this work, we focus on extrapolating model parameters to directly reduce the number of training steps and improve training efficiency.
\section{Effectiveness and Dynamics of LLM Rank-1 Subspace During RLVR}
\label{sec:empirical_study}

In this section, we first introduce the preliminaries of RLVR and low-rank representation in Section~\ref{sec:preliminary}, and then we present the empirical experiments on the rank-1 tensor of LLM parameters in Section~\ref{sec:rank_1_dominate} and Section~\ref{sec:rank_1_linearity}.
The findings drawn here serve as the motivation for our method.

\subsection{Preliminary}
\label{sec:preliminary}

\paratitle{Reinforcement learning with verifiable rewards (RLVR).}
The RLVR training dataset consists of a collection of pairs, \ie $\mathcal{D}=\{\langle x_i, a_i\rangle_{i=1}^{n}\}$, where $x_i$ and $a_i$ denote the question and the ground truth answer.
Based on question $x$, a policy with parameter $\theta$ (\ie $\pi_\theta$) will explore the solution for $G$ times, generating the solutions $\{\hat{y}_1, \dots, \hat{y}_{G}\}$.
The $i$-th solution $y_i$ contains a final answer $\hat{a}_i$.
After sampling, a verifier compares $a_i$ and $\hat{a}_i$, provide a reward $R_i$ for the $i$-th solution.
In the RLVR process, based on the question, the generated solution, and the rewards from the verifier, the parameters of the policy can be optimized.
Taking GRPO~\cite{grpo}, which is a popular RLVR algorithm, as an example, the objective function can be formulated as follows,
\begin{equation}
\small
\label{eq:grpo}
\mathcal{J}(\theta)=\mathbb{E}_{(x,y)\sim\mathcal{D},\{\hat{y}_i\}_{i=1}^{G}\sim\pi_\theta(\cdot|x)}\left[\frac{1}{G}\sum_{i=1}^{G}\frac{1}{|\hat{y}_i|}\sum_{t=1}^{|\hat{y}_i|}\min\left(r_{i,t}\hat{A}_{i,t},\text{clip}\left(r_{i,t},1-\varepsilon, 1+\varepsilon\right)\hat{A}_{i,t}-\beta D_\text{KL}\right)\right],
\end{equation}
where $r_{i,t}$ and $\hat{A}_{i,t}$ refer to the importance sampling coefficient and advantage value of the $t$-th token in the $i$-th generated solution.
Furthermore, to estimate the advantage value of each generated response, GRPO takes the model-generated response corresponding to each individual prompt as a group and performs normalization on it, which can be formulated as follows,
\begin{equation}
    A_{i,1},\dots,A_{i,|y_i|} = \frac{R_i - \text{mean}\{R_1,\dots,R_{G}\}}{\text{std}\{R_1,\dots,R_{G}\}}.
\end{equation}

\paratitle{Low-Rank Representation.}
For a floating-point matrix $W \in \mathbb{R}^{n\times m}$, it requires $n\times m$ floating-point numbers for representation, whose features are difficult to extract and analyze.
To alleviate this issue, the Singular value decomposition (SVD) algorithm~\cite{svd} decomposes the matrix $W$ into the product of three matrices, as shown below,
\begin{equation}
\mathcal{W} = \mathcal{U} \Sigma \mathcal{V}^{\top}=\sum_{i=1}^{r}\bm{\sigma}_i \bm{u}_i \bm{v}_i^\top, (\mathcal{W}\in\mathbb{R}^{m\times n},\bm{\sigma}_i\in\mathbb{R}, \bm{u}_i\in\mathbb{R}^{m\times 1}, \bm{v}_i\in\mathbb{R}^{n\times1}),
\end{equation}
where $r$ is the rank of matrix $W$, $\sigma_i$ is the $i$-th value on the diagonal of matrix $\Sigma$, and $u_i$ and $v_i$ denote the $i$-th column vectors of $U$ and $V$, respectively.
Since $\sigma_i$ is non-negative and both $u_i$ and $v_i$ are unit vectors, a larger value of $\sigma_i$ indicates that the corresponding singular vectors have a greater influence on the matrix $W$.
Therefore, assuming that $\sigma_1 > \dots > \sigma_r$, the subspace spanned by the singular vectors corresponding to $\sigma_1$, which is the largest value among $\{\sigma_1,\dots,\sigma_r\}$, has the greatest influence on matrix $W$, \ie $W_1=\sigma_1 u_1 v_1^\top$.
Following the naming convention in prior work~\cite{alpha_rl}, we refer to this subspace as the Rank-1 Subspace.

\begin{figure}[!t]
\centering
\begin{minipage}[c]{0.48\textwidth}
    \centering
    \includegraphics[width=\linewidth]{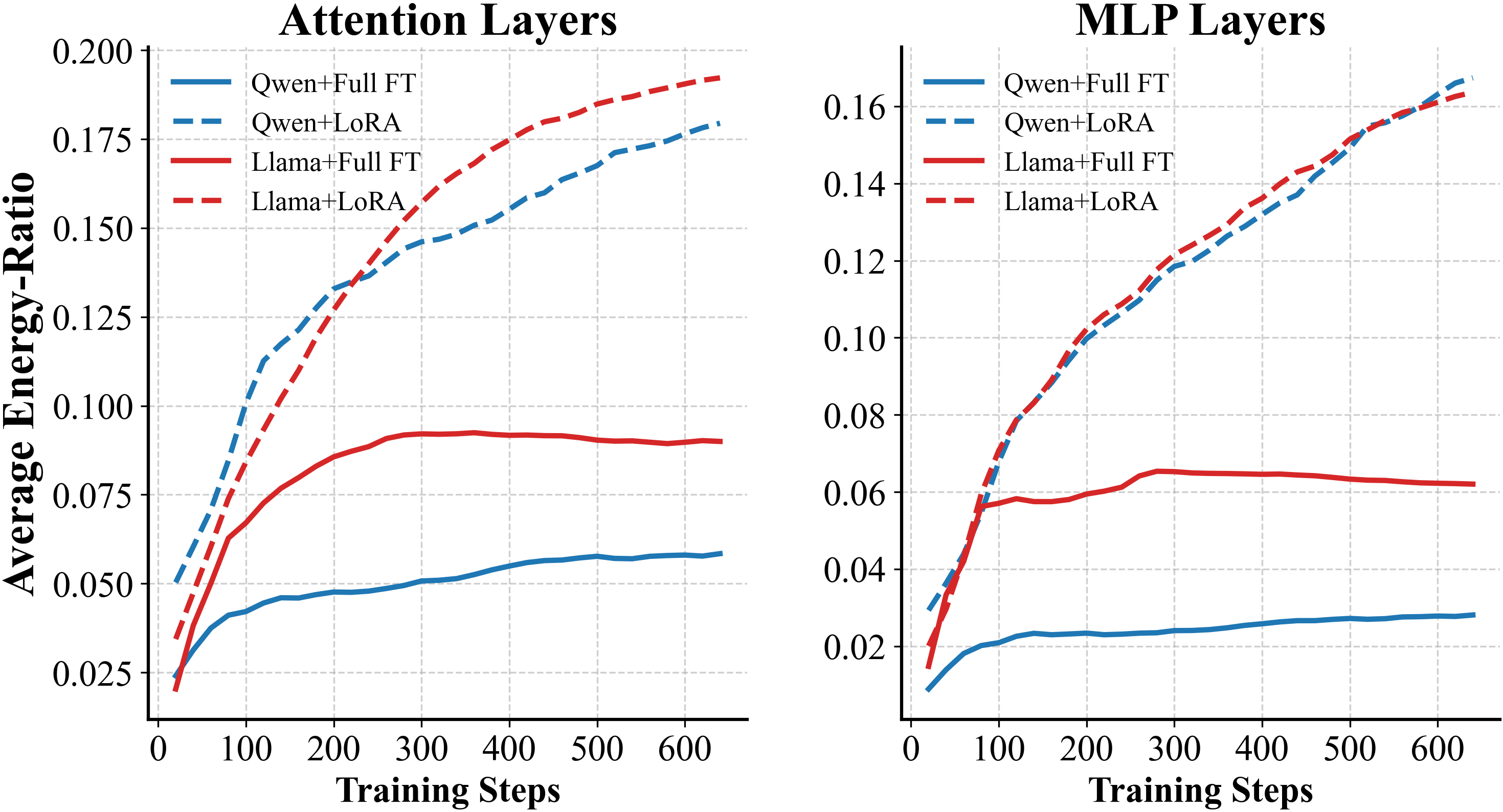}
\end{minipage}
\hspace{0.02\textwidth}
\begin{minipage}[c]{0.48\textwidth}
    \centering
    \includegraphics[width=\linewidth]{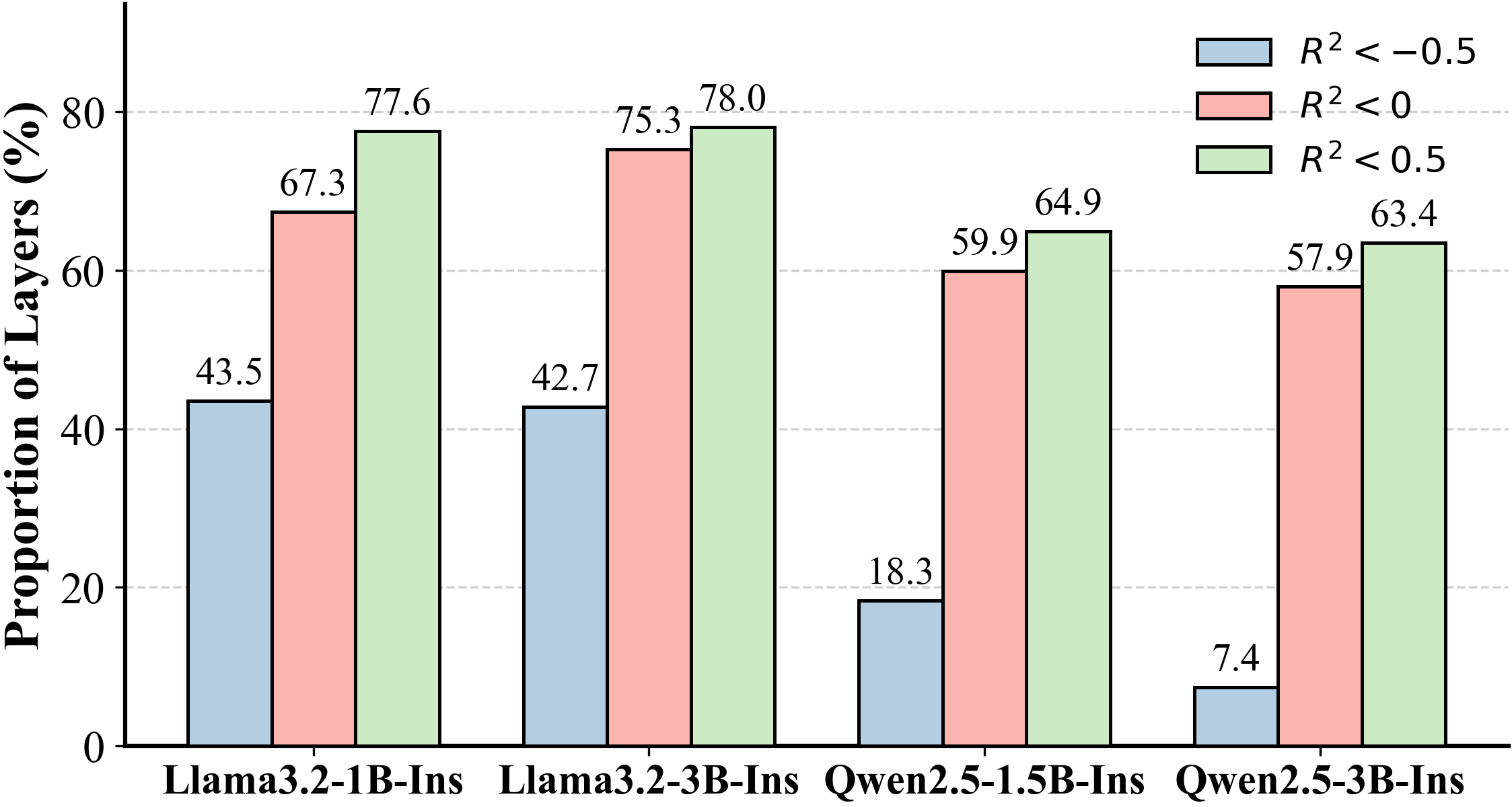}
\end{minipage}\\[3mm]
\begin{minipage}[t]{0.48\textwidth}
    \centering
    \caption{Variation of energy ratio during RLVR.}
    \label{fig:energy_ratio}
\end{minipage}
\hspace{0.02\textwidth}
\begin{minipage}[t]{0.48\textwidth}
    \centering
    \caption{R$^2$ of linear prediction across various models.}
    \label{fig:r_2}
\end{minipage}
\end{figure}



\subsection{Rank-1 Subspace Dominates LLM Parameter Updates Through RLVR}
\label{sec:rank_1_dominate}

Prior work~\cite{alpha_rl} has shown that for the parameter update matrix $\Delta W$ obtained from RLVR training, approximating it using its rank-1 subspace can largely recover the model’s performance.
In fact, the underlying reason why the rank-1 subspace can achieve such effectiveness during training remains unclear.
To investigate this issue, we examine the energy ratio~\cite{svd} of the rank-1 subspace of parameter updates $\Delta W$ throughout the RLVR process.
Formally, the energy ratio of the rank-1 subspace can be formulated as $E_1 = \sigma_1 / (\sum_{i=1}^{r}\sigma_i)$, where $\sigma_1$ denotes the largest singular value obtained from the SVD of $\Delta W$.
Moreover, inspired by low-rank approximation, we consider whether using LoRA~\cite{lora} for model fine-tuning will affect the dominance of the rank-1 subspace.

Based on the above discussion, we analyze the changes in the energy ratio of the rank-1 subspace of parameter updates under full-parameter fine-tuning (denoted as ``Full FT'') and LoRA fine-tuning, and present the results in Figure~\ref{fig:energy_ratio}.
From the experiments, we observe that in the early stage of training, the energy ratio of the rank-1 subspace in the parameter updates gradually increases, indicating that its influence on the original matrix becomes stronger, which demonstrates the reliability of using the rank-1 subspace for approximation.
Notably, models trained with LoRA exhibit a more pronounced dominance of the rank-1 subspace compared to those trained with full-parameter fine-tuning.
As shown by the dashed lines in Figure~\ref{fig:energy_ratio}, the energy ratios of the Qwen and LLaMA models grow to relatively high levels and continue to increase.
Based on this observation, we believe that LoRA training leads to a more dominant rank-1 subspace, which in turn enables better extrapolation.



\subsection{Not all Parameters in LLM RLVR satisfy Linearity}
\label{sec:rank_1_linearity}

To design a better prediction method, we examine the linearity of model parameter updates $\Delta W$.
Based on the first 10 checkpoints during RLVR, we use least-squares regression to predict the rank-1 subspace of the parameter updates for the next 5 checkpoints and compute the proportion of parameters whose R$^2$ values fall within different ranges.
We conduct experiments on four models of different sizes, and present the results in Figure~\ref{fig:r_2}.
The experimental results show that more than 50\% of the parameter updates are not well predicted, \ie R$^2$ < 0.
For a subset of parameters, the R$^2$ correlation between the predicted and true values is less than ($-0.5$).
This observation indicates that not all parameter updates within the model are linear. As a result, linear prediction methods fail to accurately capture parameter changes in the later stages of training, which may lead to degraded extrapolation performance.


\section{Extrapolation with Low-Rank Parameter Optimization Trajectories}
\label{sec:approach}

\begin{figure*}[t]
    \centering
    \includegraphics[width=\linewidth]{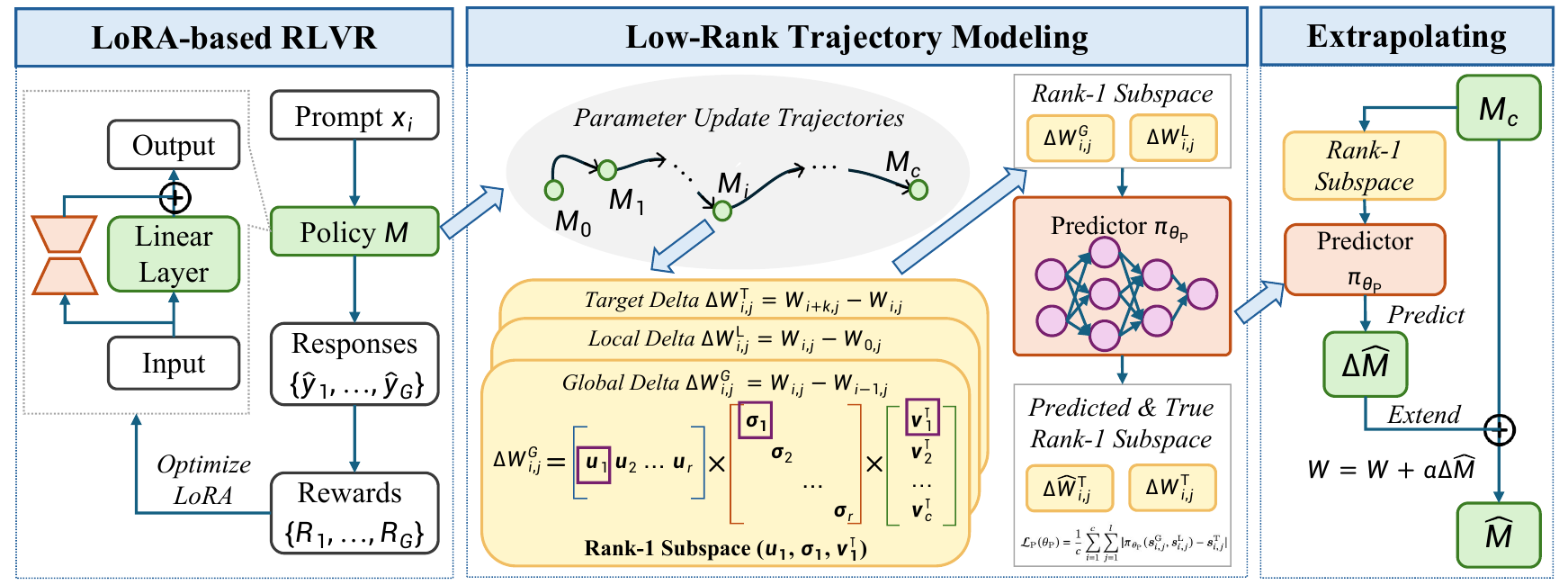}
    \caption{The overview of our NExt, containing extracting reasoning patterns (Section~\ref{sec:extract_representation}) and extrapolating model parameters (Section~\ref{sec:extrapolation}) processes. Concretely, we first utilize LoRA-based RLVR to train the LLM and save the intermediate checkpoints. Next, we extract the rank-1 subspace of the difference between these saved checkpoints, and these deltas are utilized to train the predictor. Finally, the well-trained predictor will predict the model parameters, achieving the extrapolation.}
    \label{fig:framework}
\end{figure*}

Building on the insights from our empirical experiments, in this section, we introduce \textbf{NExt}, an approach that utilizes low-rank fine-tuning and low-rank extrapolation to accelerate the RLVR process.
Concretely, we first perform RLVR training on the LLMs through LoRA fine-tuning, and then collect the reasoning representation, which is key information from the RLVR training trajectory (Section~\ref{sec:extract_representation}).
Afterward, we utilize the collected information to train a predictor $\pi_{\theta_\text{P}}$, and utilize the predictor to perform a predict-extend process to extrapolate the model parameters (Section~\ref{sec:extrapolation}).

\subsection{Low-Rank Optimization Trajectory Extracting}
\label{sec:extract_representation}

In this part, we introduce how to extract the reasoning representation from the previous RLVR trajectory. 
We first train the model for a small number of steps, and then extract the necessary information based on the parameter updates (\ie $\Delta \mathcal{W}$) before and after training.

\paratitle{LLM RLVR via LoRA Fine-tuning.}
To reduce the computational overhead incurred during subsequent extrapolation, we approximate the parameter updates using the corresponding rank-1 subspace.
As we discussed in Section~\ref{sec:rank_1_dominate}, the rank-1 subspace exhibits stronger dominance in LoRA fine-tuning and can better approximate the original parameters, thereby improving the effectiveness of subsequent extrapolation.
Building on this insight, we employ LoRA fine-tuning during the RLVR training process of the LLMs.
Concretely, we integrate the LoRA adapter into the model parameters, which can be formulated as follows,
\begin{equation}
\begin{aligned}
    \bm{h} = \mathcal{W}_ix + \Delta \mathcal{W}_ix = \mathcal{W}_i\bm{x} + \mathcal{B}_i\mathcal{A}_ix~~(\mathcal{W}_i \in \mathbb{R}^{m\times n}. \mathcal{B}_i\in \mathbb{R}^{m\times r},\mathcal{A}_i\in \mathbb{R}^{r\times n}),
\end{aligned}
\end{equation}
where $\bm{x}$ and $\bm{h}$ refer to the input and the output of the current LLM matrix, $\mathcal{B}_i$ and $\mathcal{A}_i$ are the LoRA adapters.
During LoRA fine-tuning, only the LoRA adapters (\ie $\mathcal{B}_i$ and $\mathcal{A}_i$) will be optimized, while the original parameter matrices will remain unchanged.
After LoRA training, we merge the LoRA adapters back into the original matrix to obtain the trained matrix, as shown below,
\begin{equation}
    \mathcal{W}_{i,j} = \mathcal{W}_{0,j} + \mathcal{B}_{i,j}\mathcal{A}_{i,j}
\end{equation}
where $\mathcal{W}_{i,j}$ denotes the $j$-th parameter of the $i$-th saved checkpoint, and  $\mathcal{B}_{i,j}$ and $ \mathcal{A}_{i,j}$ are the LoRA adapter of the $j$-th parameter of the $i$-th saved checkpoint.

\paratitle{Optimization Trajectory Collection.}
We save intermediate checkpoints during the LoRA-based RLVR process, which can be used to train the predictor $\pi_{\theta_\text{P}}$ for model parameter extrapolation.
Concretely, we collect both the difference between the trained model and the backbone model (referred to as the ``Global Delta'', \ie $\Delta \mathcal{W}_{i,j}^{\text{G}}$), and the difference between the current model and the model saved at the previous checkpoint (referred to as the ``Local Delta'', \ie $\Delta \mathcal{W}_{i,j}^{\text{L}}$).
The predictor $\pi_{\theta_\text{P}}$ can leverage the global delta and the local delta to extrapolate the model parameters.
Moreover, to collect the training targets for predictor training process, we also compute the difference between the current model and the model saved in the further checkpoint (referred to as the ``Target Delta'', \ie $\Delta \mathcal{W}_{i,j}^{\text{T}}$).
Formally, the definition of the global delta, the local delta, and the target delta can be formulated as follows,
\begin{equation}
\label{eq:compute_delta}
\begin{aligned}
    &\Delta \mathcal{W}_{i,j}^{\text{G}}=\mathcal{W}_{i,j} - \mathcal{W}_{0,j},\\
    &\Delta \mathcal{W}_{i,j}^{\text{L}}=\mathcal{W}_{i,j} - \mathcal{W}_{i-1,j},\\
    &\Delta \mathcal{W}_{i,j}^{\text{T}}=\mathcal{W}_{i+k,j} - \mathcal{W}_{i,j},
\end{aligned}
\end{equation}
where $k$ is a pre-specified parameter that represents the number of steps for which the model is extrapolated.
The $\pi_{\theta_\text{P}}$ learns to predict $\Delta \mathcal{W}_{i,j}^{\text{T}}$ based on the global delta $\Delta \mathcal{W}_{i,j}^{\text{G}}$ and the local delta $\Delta \mathcal{W}_{i,j}^{\text{L}}$.

\paratitle{Low-rank Approximation of Collected Trajectories.}
Since the three types of deltas we compute contain a large number of parameters, directly applying them in the training process of the predictor would incur substantial computational overhead.
Therefore, we approximate them using their rank-1 subspace to reduce the number of parameters and improve the efficiency of subsequent extrapolation.
Specifically, we perform SVD on each delta $\Delta \mathcal{W}$, obtaining the decomposed form $\Delta \mathcal{W} =\sum_{i=1}^{r}\bm{\sigma}_i \bm{u}_i \bm{v}_i$.
We select the largest singular value and its corresponding singular vectors, and multiply these three components yields the rank-1 subspace, which is an approximation to the original delta $\Delta \mathcal{W}$.
Formally, the simplification of the $\Delta \mathcal{W}_{i,j}^{\text{G}}$, $\Delta \mathcal{W}_{i,j}^{\text{L}}$, and $\Delta \mathcal{W}_{i,j}^{\text{T}}$ as be formulated as follows,
\begin{equation}
\label{eq:simplification}
    \begin{aligned}
    \Delta \mathcal{W}_{i,j}\approx \bm{\sigma}_{i,j}\cdot \bm{u}_{i,j}\cdot \bm{v}_{i,j}^\top~~(\Delta \mathcal{W}_{i,j}\in\mathbb{R}^{m\times n}, \bm{\sigma}_{i,j}\in\mathbb{R},\bm{u}_{i,j}\in\mathbb{R}^{m\times 1},\bm{v}_{i,j}\in\mathbb{R}^{n\times 1}),
    \end{aligned}
\end{equation}
where $\bm{\sigma}_{i,j}$ denotes the largest singular value of the delta $\Delta \mathcal{W}_{i,j}$, and $\bm{u}_{i,j}$ and $\bm{v}_{i,j}$ refer to the corresponding singular vectors.
We observe that after simplification, the number of parameters required to store $\Delta W$ decreases from $\mathcal{O}(n\times m)$ to $\mathcal{O}(n+m)$, significantly reducing resource overhead.

\subsection{Model Parameter Extrapolation}
\label{sec:extrapolation}

After collecting the instances that can be utilized to train the predictor $\pi_{\theta_\text{P}}$, we first introduce how to construct and train the predictor.
Next, we present the details of leveraging the trained predictor to extrapolate the model parameters through the predict-extend paradigm.

\paratitle{Constructing Trajectory Predictor.}
Our training data consists of three components, \ie the global delta, the local delta, and the target delta.
The predictor is tasked with predicting the target delta based on the global delta and the local delta.
Given that in the previous section we decomposed each delta matrix into two singular vectors and one singular value (which can be regarded as a vector belonging to $\mathbb{R}^{1\times 1}$ ), the predictor only needs to predict each vector, rather than the original delta matrix, thereby reducing computational cost.
Since $\bm{u}$, $\bm{v}$, and $\bm{\sigma}$ are predicted in the same manner, we utilize $\bm{s}^\text{G}$ to denote $\bm{u}^\text{G}$, $\bm{v}^\text{G}$, and $\bm{\sigma}^\text{G}$, in order to simplify the subsequent discussion.
Similarity, we employ $\bm{s}^{L}$ to denote $\bm{u}^\text{L}$, $\bm{v}^\text{L}$, and $\bm{\sigma}^\text{L}$; and we adopt $\bm{s}^{T}$ to denote $\bm{u}^\text{T}$, $\bm{v}^\text{T}$, and $\bm{\sigma}^\text{T}$.
Thus, the input to the predictor can be modeled as two vectors derived from the global delta and the local delta (denoted as $\bm{s}^\text{G}$ and $\bm{s}^\text{L}$), while the output is a vector derived from the target delta (denoted as $\bm{s}^\text{T}$).
Based on this, we adopt an encoder–decoder architecture to construct the predictor.
Concretely, the encoder is constructed by MLP layers and the activation function to encode the $\bm{s}^\text{G}$ and $\bm{s}^\text{L}$, and the decoder is also built by MLP layers and the activation function to decode the hidden states decode the concatenated hidden states into the target vector $\bm{s}^\text{T}$.
The follow equations can express the inference process of the predictor,
\begin{equation}
\label{eq:prediction}
    \begin{aligned}
        &\bm{h}^\text{G} = E^\text{G}(\bm{s}^\text{G}),~\bm{h}^\text{L} = E^\text{L}(\bm{s}^\text{L}),\\
        &\bm{h} = \text{Concatenate}(\bm{h}^\text{G},\bm{h}^\text{L}), \\
        &\hat{\bm{s}}^\text{T} = D(\bm{h}),
    \end{aligned}
\end{equation}
where $E^\text{G}$ and $E^\text{L}$ denote the encoder of $\bm{s}^\text{G}$ and $\bm{s}^\text{L}$, respectively, and $D$ refers to the decoder.
The illustration of the architecture of the predictor can be found in Figure~\ref{fig:framework}.

\paratitle{Modeling Optimization Trajectories.}
To train the predictor $\pi_{\theta_\text{P}}$, we constructed the training dataset, where $\bm{s}^\text{G}$ and $\bm{s}^\text{L}$ are inputs, and $\bm{s}^\text{T}$ is the target output.
Based on such training instances, our training objective is to minimize the discrepancy between the model prediction $\hat{\bm{s}}^\text{T}$ and the ground-truth target outputs $\bm{s}^\text{T}$.
As discussed in previous work~\cite{Elharrouss-arxiv-2025-Loss,Terven-ai-2025-A}, the L1 norm and L2 norm are widely adopted in model training for regression tasks.
In this work, we utilize the L1 norm for predictor optimization rather than the L2 norm, to avoid excessively small gradients brought by the L2 norm.
Formally, the objective function of the predictor training process can be shown as follows,
\begin{equation}
\label{eq:predictor_loss}
    \mathcal{L}_\text{P}(\theta_\text{P})=\frac{1}{c}\sum_{i=1}^{c}\sum_{j=1}^{l}|\pi_{\theta_\text{P}}(\bm{s}_{i,j}^\text{G},\bm{s}_{i,j}^\text{L}) - \bm{s}_{i,j}^\text{T}|,
\end{equation}
where $\pi_{\theta_\text{P}}()$ denotes the predicted vector from the predictor $\pi_{\theta_\text{P}}$, $\bm{s}_{i,j}$ denotes the singular vector of the rank-1 subspace of the $j$-th parameter in the $i$-th checkpoint, and $c$ and $l$ refer to the number of saved checkpoints and the parameters in LLMs, respectively.

\paratitle{Predict-Extend Paradigm for Extrapolation.}
To extrapolate model parameters based on training trajectories, we design the predict-extend paradigm.
First, we leverage the well-trained predictor $\pi_{\theta_\text{P}}$ to predict the model parameters in the future checkpoints.
Next, we employ an extending coefficient $\alpha$ to further extrapolation.
Concretely, for each LLM parameter $\mathcal{W}$ in the last checkpoint, we compute the global delta and local delta, and perform the SVD process on these calculated deltas.
Following Eq.~\ref{eq:prediction}, we can predict the target delta (referred to as $\Delta \hat{\mathcal{W}}^\text{T}$) based on the rank-1 subspace of $\Delta \mathcal{W}^\text{G}$ and $\Delta \mathcal{W}^\text{L}$.
Finally, we scale the predicted delta and add it back to the original parameters to complete the extrapolation, \ie
\begin{equation}
\label{eq:predict_extend}
    \hat{\mathcal{W}} = \mathcal{W} +  \alpha \cdot \Delta \hat{\mathcal{W}}.
\end{equation}

\paratitle{Improving Effectiveness and Efficiency in Practical Applications.}
To further enhance the model's extrapolation performance, we propose two optimization strategies.
First, since the singular vectors (\ie $u$ and $v$) obtained after SVD decomposition have a modulus of 1, it is necessary to normalize the singular vectors, \ie 
\begin{equation}
    \hat{\bm{s}}^\text{T}=\frac{\pi_{\theta_\text{P}}(\bm{s}_{i,j}^\text{G},\bm{s}_{i,j}^\text{L})}{|\pi_{\theta_\text{P}}(\bm{s}_{i,j}^\text{G},\bm{s}_{i,j}^\text{L})|}.
\end{equation}
Second, to improve training and prediction efficiency, we concatenate singular vectors of the same dimension and perform predictions uniformly. 
Since the predictor is constructed with MLP layers and the activation function, no interference occurs between different singular vectors.
Given the powerful parallel computing capability of GPUs, concatenating singular vectors with the same dimension accelerates both training and inference speed.

\begin{algorithm}[t]
\small
\caption{The pseudocode of the NExt algorithm.}
\label{code:NExt}
\SetKwInOut{Input}{Input}
\SetKwInOut{Output}{Output}

\Input{The backbone model $M_0$, and the RLVR training dataset $\mathcal{D}$.}
\Output{An extrapolated model $\hat{M}$.}
\BlankLine
\texttt{\# 1. Extracting Reasoning Representation.} \\
Perform RLVR training on $M_0$ through LoRA fine-tuning;\\
Collect the intermediate checkpoints during LoRA-based RLVR training, $\{M_1,\dots,M_c\}$;\\
\For{$i$ from $1$ to $c-k$}{
    \For{$j$ from $1$ to $l$}{
         Compute the deltas, including $\Delta \mathcal{W}_{i,j}^\text{G}$, $\Delta \mathcal{W}_{i,j}^\text{L}$, and $\Delta \mathcal{W}_{i,j}^\text{T}$, through Eq.~\ref{eq:compute_delta};\\
         Compute the rank-1 subspace of each delta through Eq.~\ref{eq:simplification};\\
    }
}

\BlankLine
\texttt{\# 2. Extrapolating Model Parameters.}
Initialize the predictor $\pi_{\theta_\text{P}}$ using a uniform distribution;\\
Construct the training instance for the predictor, $\mathcal{D}_\text{P}$;\\
Train the predictor $\pi_{\theta_\text{P}}$ on $\mathcal{D}_\text{P}$ under the loss function Eq.~\ref{eq:predictor_loss};\\
\For{each parameter $\mathcal{W}$ in model $M_c$}{
     Utilize the predictor $\pi_{\theta_\text{P}}$ to predict the rank-1 subspace ($\hat{\bm{\sigma}}$, $\hat{\bm{u}}$, and $\hat{\bm{v}}$) through Eq.~\ref{eq:prediction};\\
     Construct the predicted delta based on the predicted rank-1 subspace, $\Delta \hat{\mathcal{W}}=\hat{\bm{\sigma}}\cdot\hat{\bm{u}}\cdot\hat{\bm{v}}^\top$;\\
     Extrapolate the model parameter within the extending coefficient through Eq.~\ref{eq:predict_extend}; \\
}
Use the predicted parameters to construct the extrapolated model $\hat{M}$;

\BlankLine
return $\hat{M}$;
\end{algorithm}

\begin{table*}[t]
    \centering
    \small
    \setlength{\tabcolsep}{2.3pt}
    \caption{Comparison between our NExt and previous work, containing the methods for different training stages.}
      \begin{tabular}{lcccc}
      \toprule
       \textbf{Methods} & \textbf{Stage} & \textbf{Trained Param.} & \textbf{Extrapolation} & \textbf{Extrapolated Param.} \\
      \midrule
      WSM~\cite{wsm} & Pre-training & Full Parameters & Linear & Full Parameters \\
      MAEC~\cite{maec} & Pre-training & Key Neurons & Linear & Key Neurons \\
      \midrule
      DARE~\cite{yu2024language} & SFT & Full Parameters & Linear & Randomly Selected \\
      Greedy Soup~\cite{wortsman2022model} & SFT & Full Parameters & Linear & Full Parameters  \\
      \midrule
      AlphaRL~\cite{alpha_rl} & RLVR & Full Parameters & Linear & Rank-1 Subspace \\
      RL-Extra~\cite{rl_extra} & RLVR & Full Parameters & Linear & Full Parameters \\
      ExPO~\cite{zheng2025model} & Alignment & Full Parameters & Linear & Full Parameters \\
      \midrule
      \textbf{NExt (Ours)} & \textbf{RLVR} & \textbf{LoRA Adapter} & \textbf{Non-Linear} & \textbf{Rank-1 Subspace} \\
      \bottomrule
      \end{tabular}
      \label{tab:comparison}
\end{table*}

\subsection{Summarization and Discussion}

To further demonstrate the workflow of NExt, we provide pseudocode in Algorithm~\ref{code:NExt}, including reasoning representation extracting and model parameters extrapolating process of our method.
Concretely, we first perform RLVR training on the backbone model using LoRA and save the parameters of intermediate checkpoints $\{M_1,\dots,M_c\}$ during training.
Next, we compute the global delta, local delta, and target delta of each checkpoint via Eq.~\ref{eq:compute_delta}, and then calculate the rank-1 subspace of the deltas through Eq.~\ref{eq:simplification}, which are utilized to construct the training dataset of the predictor.
Afterward, we initialize the predictor $\pi_{\theta_\text{P}}$ in the uniform distribution and then leverage the constructed dataset to train $\pi_{\theta_\text{P}}$ through Eq.~\ref{eq:predictor_loss}.
Finally, obtaining a well-trained predictor, we employ it to predict the model parameters based on the last checkpoint $M_c$, and then adopt the coefficient to further extend the predicted parameters.
During the above process, we can obtain the extrapolated LLM $\hat{M}$, which can be further trained through the RLVR to improve performance.


To further highlight the distinction of our NExt, we provide a systematic comparison with representative parameter extrapolation methods in Table~\ref{tab:comparison}, covering different training stages and design choices. From the comparison, it can be observed that existing methods mainly differ along three key dimensions, including the training stage, the form, and the subset of parameters involved.
Specifically, prior works (\eg WSM, MAEC, and DARE) primarily focus on the pre-training or supervised fine-tuning (SFT) stages, where extrapolation is typically performed over full parameters or selected subsets. These methods generally rely on linear combinations of model parameters to improve model performance. In contrast, more recent methods designed for RLVR, such as AlphaRL [11] and RL-Extra [12], attempt to reduce costs by extrapolating parameters along the RLVR trajectory. However, these methods still adopt linear extrapolation strategies, either in the full parameter space or within a rank-1 subspace, implicitly assuming the linearity of parameter evolution.

Despite their effectiveness, such linear assumptions may not sufficiently capture the complex dynamics of parameter updates during RLVR training, as demonstrated by our empirical findings in Section~\ref{sec:empirical_study}. In particular, although the rank-1 subspace exhibits strong dominance, its evolution is inherently non-linear, which limits the effectiveness of linear extrapolation methods in later training stages.
In contrast to existing approaches, our NExt introduces a fundamentally different perspective by modeling the parameter optimization trajectory in a nonlinear manner. First, instead of performing extrapolation on full parameters, we leverage LoRA-based fine-tuning to induce a more structured and dominant low-rank representation, and conduct extrapolation within the rank-1 subspace to improve efficiency. Second, rather than relying on predefined linear combinations, we explicitly learn the evolution pattern of parameter updates through a predictor, enabling more flexible and accurate modeling of training dynamics. Third, our method is specifically designed for the RLVR setting, where it directly reduces the number of required training steps by predicting future parameter states, instead of solely improving training effectiveness.

Overall, NExt differs from prior work in that it bridges low-rank modeling and nonlinear trajectory prediction within the RLVR framework, providing a unified solution that improves both training efficiency and model performance. This design not only complements existing acceleration techniques, but also offers a new perspective for accelerating LLM optimization.

\section{Experiment}
\label{sec:experiment}

In this section, we introduce the details of experiment settings and the main results, and then we conduct the detailed analysis for further understanding of our approach.

\subsection{Experimental Settings}

In this part, we present the datasets, baselines, and evaluation metrics in our experiment.
We also provide the details of implementation and hyperparameters.


\begin{wraptable}{r}{0.45\columnwidth}
    \centering
    \caption{The hyperparameters used in our experiments.}
    \setlength{\tabcolsep}{2.5pt}
    \small
      \begin{tabular}{lcc}
      \toprule
       \textbf{Process} & \textbf{Hyperparameters} & \textbf{Value} \\
      \midrule
      \multirow{11}*{Train} & Train Batch Size & 128 \\
      & Mini Batch Size & 32 \\
      & Num. Rollout & 8 \\
      & Rollout Temperature & 1.0 \\
      & Rollout top\_p & 1.0 \\
      & Max Prompt Length & 1024 \\
      & Max Response Length & 4096 \\
      & Learning Rate For FP & $5\times 10^{-7}$ \\
      & Learning Rate For LoRA & $5\times 10^{-6}$ \\
      & LoRA Rank & 64 \\
      & LoRA Alpha & 32 \\
      \midrule
      \multirow{4}*{Test} & Max Response Length & 4096 \\
      & temperature & 1.0 \\
      & top\_p & 1.0 \\
      & Number of Repeated Runs & 8 \\
      \bottomrule
      \end{tabular}
      \label{tab:hyperparameters}
\end{wraptable}

\paratitle{Datasets.}
For RLVR, we adopt the dataset in the previous work~\cite{dapo}, which contains approximately $17$k mathematical reasoning problems.
For the evaluation process, we conduct the five mathematical tasks, \ie AIME24~\cite{aime24}, AIME25~\cite{aime25}, AMC23~\cite{amc23}, Minerva~\cite{minerva}, and the easy version of OlymMATH~\cite{olymmath}.
These tasks cover a range of difficulty levels and also include challenging competition problems.

\paratitle{Baselines.}
First, we take GRPO~\cite{grpo} with full-parameter fine-tuning (\ie GRPO w/ FP) and LoRA fine-tuning (\ie GRPO w/ LoRA)~\cite{lora} as baseline methods to compare the performance changes of the model after applying NExt acceleration.
Next, among existing acceleration methods, AlphaRL~\cite{alpha_rl} and RL-Extra~\cite{rl_extra} have been used to improve the training efficiency of RLVR, and we also include them as baselines for comparison.

\paratitle{Implementation Details.}
To better help readers understand and apply our method, we present the hyperparameters of the RLVR process for our experiment in Table~\ref{tab:hyperparameters}.
Besides, we employ GRPO~\cite{grpo} as the backbone RLVR algorithm in our experiments.
For the hyperparameters of our NExt, we set the extending coefficient $\alpha$ as $1.5$ to extrapolate the predicted parameters.
Moreover, during the first $150$ steps of RLVR, we save a checkpoint every $10$ steps, resulting in a total of $15$ checkpoints for subsequent extrapolation. 
We set $k=5$, indicating that the predictor $\pi_{\theta_\text{P}}$ is trained to predict the outcome after 50 training steps.
After extrapolation, we further conduct 100 additional RLVR training steps to improve model performance.
In summary, we perform a total of 250 RLVR steps during the NExt.

\paratitle{Evaluation Metrics}
In our experiments, we report the number of training steps and the average accuracy of the trained model across different tasks.
To ensure the stability of the experimental results, we repeat each task eight times and use the average accuracy as the model’s final performance on that task.
Additionally, to more intuitively demonstrate the acceleration achieved by the algorithm, we report the \emph{the incremental cost-effectiveness ratio (ICER)}~\cite{ICER}, \ie $\texttt{ICER} = \#\texttt{{Step}} / \texttt{Improvement} \times 100\%$, where lower values indicate higher efficiency of the algorithm that can utilize less resources to achieve better performance.

\begin{table*}[t]
    \centering
    \small
    \setlength{\tabcolsep}{3.5pt}
    \caption{Accuracy of LLMs with larger than 7B parameters trained through different methods on mathematical tasks. The best is in bold.}
      \begin{tabular}{lcccccccc}
      \toprule
       \textbf{Methods} & \textbf{\#Steps} & \textbf{AIME24} & \textbf{AIME25} & \textbf{AMC23} & \textbf{Minerva} & \textbf{OlymMATH} & \textbf{Avg.} & \textbf{ICER} ($\downarrow$) \\
      \midrule
      \multicolumn{9}{c}{\textit{Qwen2.5-7B-Instruct}} \\
      Backbone Model & - &  10.0 & 5.4 & 51.9 & 22.9 & 5.3 & 19.1   & - \\
      + GRPO w/ FP   & 250 & 13.8 & 11.7 & 59.1 & 24.9 & 6.0 & 23.1 & 62.5 \\
      + GRPO w/ LoRA & 250 & 13.3 & 10.8 & 56.3 & 24.9 & 5.0 & 22.1 & 83.3 \\
      + GRPO w/ FP   & 400 & 16.3 & 11.3 & 59.7 & 25.9 & 6.8 & 24.0  & 81.6 \\
      + GRPO w/ LoRA & 400 & 16.7 & 11.7 & 59.4 & 24.7 & 5.0 & 23.5 & 90.9 \\
      + AlphaRL      & 250 & 14.6 & 8.8 & 55.9 & 23.8 & 5.0 & 21.6 & 100.0 \\
      + RL-Extra     & 250 & 15.4 & 8.8 & 58.8 & 24.9 & 5.8 & 22.7 & 69.4 \\
      + NExt (Ours)  & 250 & 16.4 & 12.5 & 60.3 & 25.5 & 6.5 & \textbf{24.2} & \textbf{49.0} \\
      \midrule
      \multicolumn{9}{c}{\textit{Qwen2.5-14B-Instruct}} \\
      Backbone Model & - & 12.1 & 9.2 & 51.3 & 26.1 & 5.4 & 20.8  & - \\
      + GRPO w/ FP   & 250 & 13.3 & 14.2 & 65.9 & 29.4 & 8.8 & 26.3 & 45.5 \\
      + GRPO w/ LoRA & 250 & 14.2 & 15.0 & 62.5 & 29.7 & 7.8 & 25.8 & 50.0 \\
      + GRPO w/ FP   & 400 & 17.1 & 17.5 & 66.3 & 29.0 & 8.8 & 27.7 & 58.0 \\
      + GRPO w/ LoRA & 400 & 16.7 & 13.3 & 65.0 & 31.3 & 8.8 & 27.0 & 64.5 \\
      + AlphaRL      & 250 & 13.3 & 12.1 & 63.4 & 28.6 & 7.8 & 25.0 & 59.5 \\
      + RL-Extra     & 250 & 15.4 & 14.6 & 63.1 & 30.2 & 7.6 & 26.2 & 46.3 \\
      + NExt (Ours)  & 250 & 17.9 & 16.7 & 67.2 & 30.5 & 9.3 & \textbf{28.3} & \textbf{33.3} \\
      \bottomrule
      \end{tabular}
      \label{tab:main_results_large}
\end{table*}

\begin{table}[t]
    \centering
    \small
    \setlength{\tabcolsep}{3.5pt}
    \caption{Accuracy of LLMs with fewer than 3B parameters trained through different methods on mathematical tasks. The best is in bold.}
      \begin{tabular}{lccccccccc}
      \toprule
       \multirow{2.5}*{\textbf{Methods}} & \multirow{2.5}*{\textbf{\#Steps}} & \multicolumn{4}{c}{\textbf{Qwen2.5-1.5B-Instruct}} & \multicolumn{4}{c}{\textbf{Qwen2.5-3B-Instruct}} \\
       \cmidrule(r){3-6} \cmidrule(r){7-10}
       & & \textbf{AMC23} & \textbf{Minerva} & \textbf{Avg.} & \textbf{ICER} ($\downarrow$) & \textbf{AMC23} & \textbf{Minerva} & \textbf{Avg.} & \textbf{ICER} ($\downarrow$) \\
      \midrule
      Backbone Model & - & 16.3 & 7.4 & 11.9 & -& 31.3 & 15.7 & 23.5 & -  \\
      + GRPO \textit{w/} FP   & 250 & 26.7 & 11.1 & 18.9 & 35.7 & 40.6 & 18.3 & 29.5 & 41.7 \\
      + GRPO \textit{w/} LoRA & 250 & 29.4 & 11.2 & 20.3 & 29.8 & 36.9 & 17.6 & 27.3 & 65.8 \\
      + GRPO \textit{w/} FP   & 400 & 31.3 & 10.8 & 21.1 & 43.5 & 42.5 & 18.8 & 30.7 & 55.6 \\
      + GRPO \textit{w/} LoRA & 400 & 30.0 & 11.5 & 20.8 & 44.9 & 40.0 & 18.8 & 29.4 & 67.8 \\
      + AlphaRL      & 250 & 27.5 & 11.8 & 19.7 & 32.1 & 39.4 & 17.3 & 28.4 & 51.0 \\
      + RL-Extra     & 250 & 26.3 & 11.1 & 18.7 & 36.8 & 41.3 & 17.9 & 29.6 & 41.0 \\
      + NExt (Ours)  & 250 & 31.3 & 11.8 & \textbf{21.6} & \textbf{25.8} & 43.1 & 18.8 & \textbf{31.0} & \textbf{33.3} \\
      \bottomrule
      \end{tabular}
      \label{tab:main_results_small}
\end{table}

\subsection{Main Results}

To make the experiments more convincing, we conduct experiments on four models of different scales on five challenging tasks, and present the results in Table~\ref{tab:main_results_large} and Table~\ref{tab:main_results_small}.


First, we can observe that NExt requires fewer training steps to achieve higher performance compared to both full-parameter fine-tuning and LoRA fine-tuning of the RLVR process.
In our experiments, we train the model for 150 steps and then extrapolate the parameters based on the obtained checkpoin, which is further trained with RLVR. 
With only 250 additional training steps, it surpasses the performance of vanilla RLVR trained for 400 steps on all four scale LLMs.
These results demonstrate that NExt can effectively accelerate the RLVR training process.

Second, compared with powerful baseline methods (\ie AlphaRL and Rl-Extra), NExt achieves better performance under the same number of training steps, with higher accuracy and lower ICER.
As shown in the experiments and analysis in Section~\ref{sec:rank_1_linearity}, not all parameters in LLMs change linearly.
Baseline methods use linear prediction to extrapolate model parameters, which may cause the parameters to deviate from the optimal optimization direction, thereby limiting improvements in model capability.
In contrast, NExt employs a more effective prediction algorithm, placing the model parameters in a better state during extrapolation, thereby providing greater learning potential in subsequent RLVR training.

Third, according to the experiment results, the training performance of LoRA fine-tuning is comparable to that of full-parameter fine-tuning.
We report the model performance of the RLVR process after 250 and 400 training steps. Across models of different scales, the two training settings yield comparable performance, \eg 21.1\% \textit{v.s} 20.8\% for 1.5B LLM, and 24.0\% \textit{v.s} 23.5\% for 7B LLM.
These results indicate that, in the RLVR training process, LoRA can partially replace full-parameter fine-tuning, achieving comparable performance with fewer resources, which reduces the training cost and enables further scaling.

\subsection{Detailed Analysis}

To further understand the feature of NExt, we conduct a detailed analysis, including an ablation study, a discussion about the consumption of computational resources, and an analysis of the extending process.
Moreover, we present further analysis about the adaptation of other RLVR algorithms and other domain tasks.

\subsubsection{Ablation Study}
To assess the effectiveness of each module in our approach, we conduct an ablation study on LLMs of different scales and present the results in Table~\ref{tab:ablation_study}.
We provide the results obtained from different training methods after extrapolation, and also report the performance of the extrapolated models after further RLVR training.
First, we find that extrapolating based on models trained with full-parameter fine-tuning (\ie ``\textit{w/o} LoRA'') performs worse than using models fine-tuned with LoRA.
This experimental observation further corroborates the conclusion in Section~\ref{sec:rank_1_dominate}.
During LoRA fine-tuning, the rank-1 subspace is better maintained in a dominant position, thereby reducing the approximation error.
Moreover, when we remove either the global delta or the local delta before parameter extrapolation, the LLM performance degrades, and subsequent RLVR training fails to recover it to a high level, indicating that both the global delta and local delta are crucial in the extrapolation. 
Leveraging information at different granularities enables better estimation of the direction and trend of future parameter updates in LLMs, leading to more significant acceleration.


\begin{wrapfigure}{r}{0.5\textwidth}
    \includegraphics[width=\linewidth]{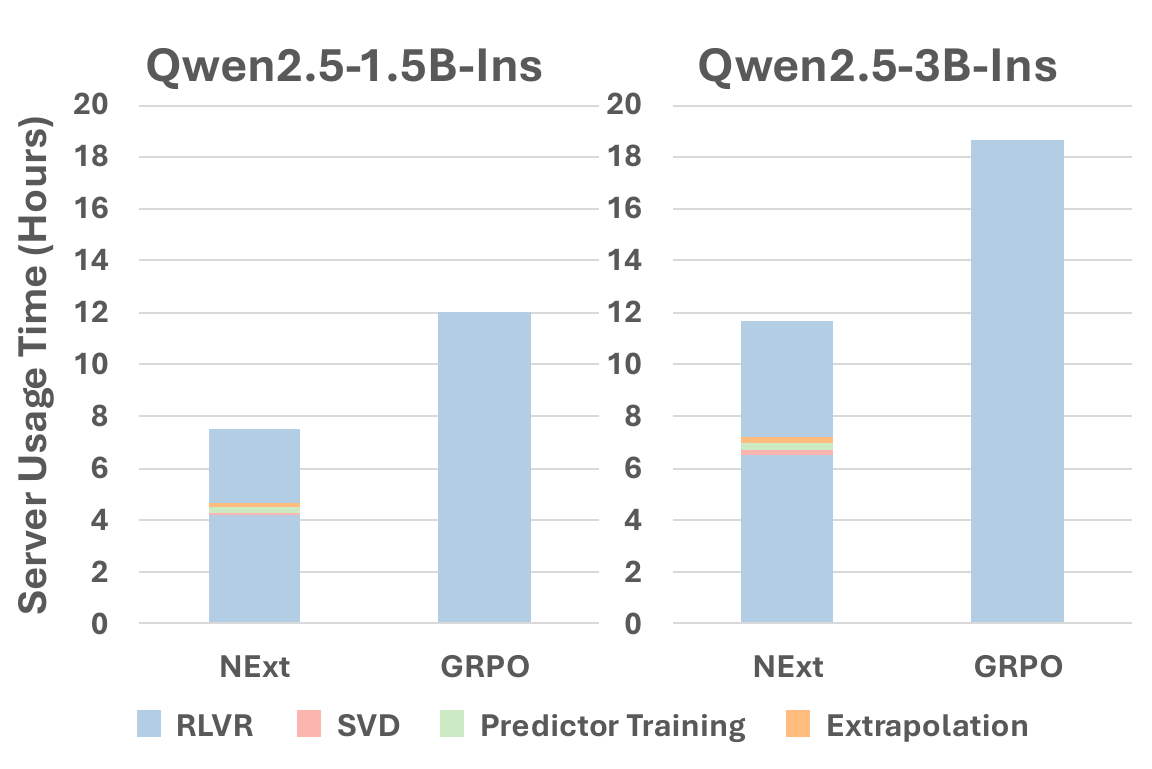}
    \caption{The comparison of server usage between NExt and GRPO.}
    \label{fig:resource_comsumption}
\end{wrapfigure}

\subsubsection{Consumption of Computation Resources}
To further analyze the resource consumption of our proposed method NExt, we measure the running time of NExt and GRPO on a 4$\times$A800 server, and present the results in Figure~\ref{fig:resource_comsumption}.
First, we observe that NExt requires significantly less server time than GRPO, \ie the training time is reduced from 18.7 hours to 11.7 hours for the 3B model and reduced from 12 hours to 7.4 hours for the 1.5B model, achieving a 37.5\% speedup.
This result demonstrates that NExt can successfully accelerate the RLVR process, thereby resulting in reduced resource overhead.
Furthermore, the newly introduced processes in NExt, including SVD, predictor training, and extrapolation, account for only a very small proportion of the overall training cost.
This observation indicates that NExt not only effectively reduces the number of training steps in RLVR, achieving better performance with fewer steps, but also does not introduce significant time overhead, thereby reducing overall server usage time.

\begin{table*}[t]
    \centering
    \small
    \caption{Ablation study of our NExt. ``\textit{w/o} LoRA'' denotes that the RLVR process optimizes full parameters in LLMs. ``\textit{w/o} G-Delta'' and ``\textit{w/o} L-Delta'' refer to extrapolation without global delta and local delta, respectively.}
      \begin{tabular}{lcccccccc}
      \toprule
       \multirow{2.5}*{\textbf{Methods}} & \multicolumn{4}{c}{\textbf{Predict-Extend Process}} & \multicolumn{4}{c}{\textbf{RLVR after Extrapolation}} \\
       \cmidrule(r){2-5} \cmidrule(r){6-9}
       & \textbf{AMC23} & \textbf{Minerva} & \textbf{Avg.} & \textbf{ICER} ($\downarrow$) & \textbf{AMC23} & \textbf{Minerva} & \textbf{Avg.} & \textbf{ICER} ($\downarrow$) \\
      \midrule
      \multicolumn{9}{c}{\textit{Qwen2.5-1.5B-Instruct}} \\
      NExt (Ours) & 26.3 & 11.2 & \textbf{18.8} & \textbf{21.7} & 31.3 & 11.8 & \textbf{21.6} & \textbf{25.8}  \\
      \quad \textit{w/o} LoRA & 23.8 & 11.4 & 17.6 & 26.3 & 28.1 & 11.5 & 19.8 & 31.6  \\
      \quad \textit{w/o} G-Delta & 21.3 & 10.4 & 15.9 & 37.5 & 28.8 & 10.8 & 19.8 & 31.6  \\
      \quad \textit{w/o} L-Delta & 23.1 & 9.9 & 16.5 & 32.6 & 26.9 & 10.2 & 18.6 & 37.3  \\
      \midrule
      \multicolumn{9}{c}{\textit{Qwen2.5-3B-Instruct}} \\
      NExt (Ours) & 40.0 & 18.1 & \textbf{29.1} & \textbf{26.8} & 43.1 & 18.8 & \textbf{31.0} & \textbf{33.3}  \\
      \quad \textit{w/o} LoRA & 39.4 & 17.2 & 28.3 & 31.3 & 38.8 & 18.6 & 28.7 & 48.1  \\
      \quad \textit{w/o} G-Delta. & 38.1 & 16.6 & 27.4 & 38.5 & 38.8 & 17.6 & 28.2 & 53.2  \\
      \quad \textit{w/o} L-Delta & 36.9 & 18.1 & 27.5 & 37.5 & 40.6 & 16.9 & 28.8 & 47.2  \\
      \bottomrule
      \end{tabular}
      \label{tab:ablation_study}
\end{table*}

\begin{figure*}
    \centering
    \includegraphics[width=\linewidth]{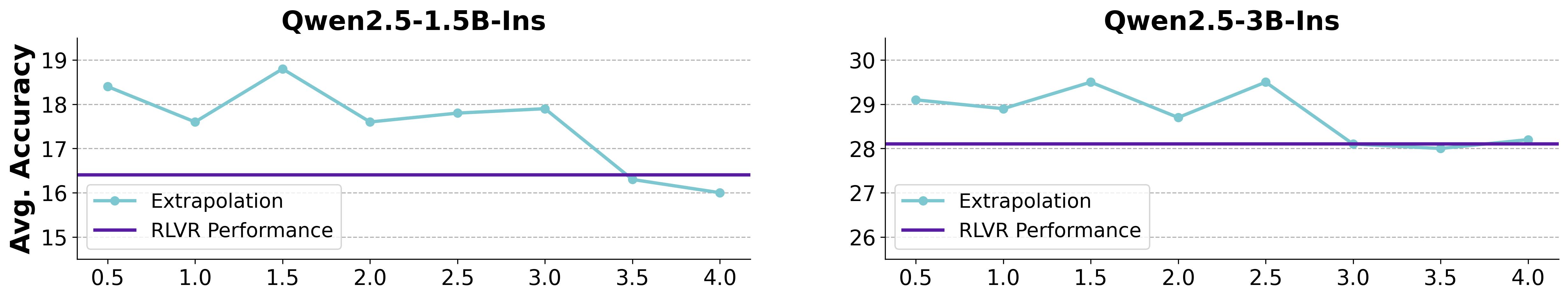}
    \caption{Performance on mathematical tasks as the extending coefficient $\alpha$ varies from 0.5 to 4.0.}
    \label{fig:ratio}
\end{figure*}

\subsubsection{Effect of Extending Process}
To evaluate the robustness of our algorithm, we conduct experiments on different values of the extending coefficient $\alpha$. 
We select eight values ranging from 0.5 to 4.0 for parameter extrapolation, and plot the curve of the model’s average performance in Figure~\ref{fig:ratio}.
From the experimental results, we observe that the model performance remains relatively stable as $\alpha$ varies.
When $\alpha \in [0.5,2.5]$, the model consistently achieves better performance than before extrapolation, indicating that our method is not highly sensitive to hyperparameter settings.
When the extending coefficient becomes too large, the model performance exhibits significant fluctuations.
This phenomenon also suggests that linearly predicting model parameters can be unstable, which is consistent with Section~\ref{sec:rank_1_linearity}.
When the extending coefficient becomes too large, the model performance is difficult to further improve and may even degrade, indicating the limitations of linear extrapolation. The predict–extend process can help mitigate this issue.
In conclusion, the results shows the effectiveness and robustness of NExt.

\begin{table}[t]
    \centering
    \small
    \caption{Performance of NExt adapted to different RLVR algorithms.}
      \begin{tabular}{lccccccc}
      \toprule
       \multirow{2.5}*{\textbf{Methods}} & \multirow{2.5}*{\textbf{\#Steps}} & \multicolumn{3}{c}{\textbf{Qwen2.5-1.5B-Instruct}} & \multicolumn{3}{c}{\textbf{Qwen2.5-3B-Instruct}} \\
       \cmidrule(r){3-5} \cmidrule(r){6-8}
       & & \textbf{AMC23} & \textbf{Minerva} & \textbf{Avg.} & \textbf{AMC23} & \textbf{Minerva} & \textbf{Avg.} \\
      \midrule
      Backbone Model & - & 16.3 & 7.4 & 11.9 & 31.3 & 15.7 & 23.5 \\
      \midrule
      + RLOO & 250 & 20.0 & 7.4 & 13.7 & 34.4 & 16.3 & 25.4  \\
      + RLOO & 400 & 22.5 & 8.3 & 15.4 & 37.5 & 17.0 & 27.3  \\
      + RLOO \textit{w}/ NExt & 250 & 24.4 & 10.6 & 17.5 & 38.4 & 18.5 & 28.5  \\
      \midrule
      + REINFORCE++ & 250 & 17.5 & 7.8 & 12.7 & 33.1 & 16.3 & 24.7  \\
      + REINFORCE++ & 400 & 21.9 & 9.3 & 15.6 & 36.3 & 16.6 & 26.5  \\
      + REINFORCE++ \textit{w}/ NExt & 250 & 22.5 & 9.0 & 15.8 & 38.8 & 17.0 & 27.9  \\
      \bottomrule
      \end{tabular}
      \label{tab:other_rlvr_algorithms}
\end{table}

\subsubsection{Adaptation of Other RLVR Algorithms}

Since NExt is not designed to exploit the characteristics of any specific RLVR method, our acceleration method is orthogonal to RLVR algorithms in principle, \ie our method can be applied to any RLVR algorithm.
To validate this assumption, we apply NExt to different algorithms (\ie RLOO~\cite{rloo} and REINFORCE++~\cite{reinforce++}) and compare the performance differences between models trained with traditional RLVR and those trained with NExt acceleration.
We present the results in Table~\ref{tab:other_rlvr_algorithms}.

From the experimental results, we observe that by scaling the number of RLVR training steps, both RLOO and REINFORCE++ can improve the performance of the backbone model.
Despite differences in the training algorithms, NExt leads to better performance of LLMs than vanilla RLVR according to similar training steps.
Moreover, our NExt reduces the number of training steps by 37.5\% while enabling the resulting models to achieve comparable or even better performance.
This observation indicates that NExt does not depend on any specific RLVR algorithm, demonstrating strong generalization ability and the capability to adapt to a variety of training methods.

\begin{figure*}[t]
    \centering
    \includegraphics[width=\linewidth]{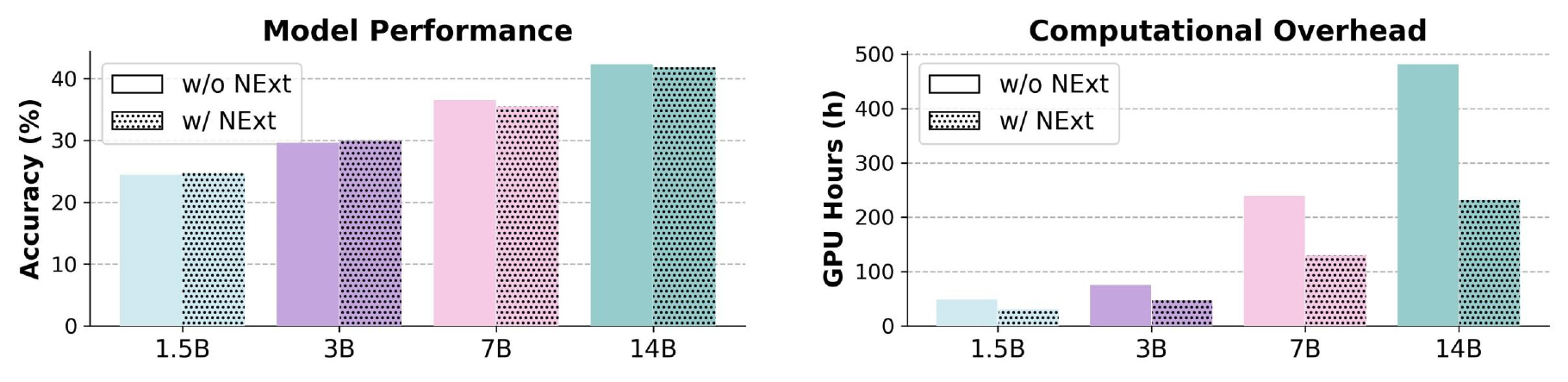}
    \caption{Comparison of performance and computational cost of LLMs trained through different methods on the GPQA task.}
    \label{fig:gpqa}
\end{figure*}

\begin{table*}[t]
    \centering
    \small
    \caption{Accuracy of LLMs at different scales on the MMLU-Pro sub-benchmarks (Part 1). We perform 400 training steps on GRPO and 250 training steps for other methods.}
    \setlength{\tabcolsep}{4pt}
      \begin{tabular}{lcccccccc}
      \toprule
       \textbf{Methods} & \textbf{Biology} & \textbf{Business} & \textbf{Chemistry} & \textbf{CS} & \textbf{Economics} & \textbf{Engineering} & \textbf{Health} & \textbf{Avg.} \\
      \midrule
      \multicolumn{9}{c}{\textit{Qwen2.5-1.5B-Instruct}} \\
      Backbone    & 33.3 & 23.0 & 14.9 & 14.6 & 29.8 & 11.6 & 19.1 & 20.9 \\
      + GRPO      & 48.3 & 39.8 & 22.1 & 27.4 & 40.5 & 13.2 & 26.4 & \textbf{31.1} \\
      + AlphaRL   & 44.2 & 38.2 & 21.7 & 21.1 & 41.7 & 15.9 & 27.9 & 30.1 \\
      + RL-Extra  & 42.1 & 35.1 & 20.5 & 23.8 & 42.1 & 10.2 & 24.1 & 28.3 \\
      + NExt (Ours)     & 46.8 & 40.1 & 22.7 & 28.7 & 39.8 & 13.2 & 25.7 & \underline{31.0} \\
      \midrule
      \multicolumn{9}{c}{\textit{Qwen2.5-3B-Instruct}} \\
      Backbone    & 65.3 & 49.6 & 35.7 & 39.0 & 49.2 & 25.3 & 43.7 & 44.0 \\
      + GRPO      & 73.6 & 54.4 & 41.9 & 41.5 & 51.0 & 27.0 & 48.0 & \underline{48.2} \\
      + AlphaRL   & 67.7 & 53.7 & 41.8 & 40.9 & 50.3 & 26.8 & 45.1 & 46.6 \\
      + RL-Extra  & 70.3 & 51.4 & 37.4 & 42.7 & 49.1 & 30.2 & 42.8 & 46.3 \\
      + NExt (Ours)      & 72.9 & 54.7 & 42.6 & 43.8 & 50.5 & 27.7 & 49.1 & \textbf{48.8} \\
      \midrule
      \multicolumn{9}{c}{\textit{Qwen2.5-7B-Instruct}} \\
      Backbone    & 77.3 & 65.0 & 48.4 & 50.3 & 63.2 & 35.3 & 59.1 & 56.9 \\
      + GRPO      & 76.7 & 65.5 & 48.2 & 52.1 & 67.6 & 39.5 & 59.1 & \underline{58.4} \\
      + AlphaRL   & 76.6 & 66.4 & 50.2 & 53.7 & 66.9 & 32.4 & 61.4 & 58.2 \\
      + RL-Extra  & 76.7 & 63.7 & 50.3 & 52.4 & 65.1 & 36.7 & 59.9 & 57.8 \\
      + NExt (Ours)      & 76.9 & 66.8 & 49.5 & 53.6 & 68.1 & 37.2 & 60.0 & \textbf{58.9} \\
      \midrule
      \multicolumn{9}{c}{\textit{Qwen2.5-14B-Instruct}} \\
      Backbone    & 80.4 & 72.0 & 61.1 & 63.7 & 70.9 & 48.1 & 66.8 & 66.1 \\
      + GRPO      & 82.8 & 79.0 & 63.4 & 72.5 & 73.4 & 52.5 & 69.7 & \textbf{70.5} \\
      + AlphaRL   & 82.8 & 76.7 & 63.9 & 70.7 & 72.9 & 51.5 & 66.8 & 69.3 \\
      + RL-Extra  & 81.1 & 80.0 & 63.3 & 69.8 & 73.0 & 50.6 & 68.8 & 69.5 \\
      + NExt (Ours)      & 82.6 & 78.2 & 64.3 & 73.0 & 73.2 & 52.7 & 69.5 & \textbf{70.5} \\
      \bottomrule
      \end{tabular}
      \label{tab:mmlu_pro_1}
\end{table*}

\begin{table*}[t]
    \centering
    \small
    \caption{Accuracy of LLMs at different scales on the MMLU-Pro sub-benchmarks (Part 2). We perform 400 training steps on GRPO and 250 training steps for other methods.}
      \begin{tabular}{lcccccccc}
      \toprule
       \textbf{Methods} & \textbf{History} & \textbf{Law} & \textbf{Math} & \textbf{Philosophy} & \textbf{Physics} & \textbf{Psychology} & \textbf{Other} & \textbf{Avg.} \\
      \midrule
      \multicolumn{9}{c}{\textit{Qwen2.5-1.5B-Instruct}} \\
      Backbone    & 16.2 &  8.5 & 29.1 & 14.5 & 16.1 & 21.7 & 18.0 & 17.7 \\
      + GRPO      & 18.8 & 19.4 & 44.7 & 19.8 & 21.2 & 39.8 & 26.4 & \underline{27.2} \\
      + AlphaRL   & 16.1 & 17.0 & 40.6 & 20.9 & 18.0 & 40.3 & 26.7 & 25.7 \\
      + RL-Extra  & 19.3 & 16.5 & 38.7 & 16.9 & 20.7 & 38.8 & 27.2 & 25.4 \\
      + NExt (Ours)      & 17.6 & 18.3 & 45.2 & 20.9 & 22.0 & 41.1 & 26.1 & \textbf{27.3} \\
      \midrule
      \multicolumn{9}{c}{\textit{Qwen2.5-3B-Instruct}} \\
      Backbone    & 32.4 & 18.3 & 55.0 & 35.7 & 36.0 & 51.9 & 37.5 & 38.1 \\
      + GRPO      & 34.1 & 20.5 & 60.6 & 35.5 & 41.7 & 54.2 & 36.8 & \textbf{40.5} \\
      + AlphaRL   & 32.4 & 19.6 & 60.8 & 33.0 & 41.9 & 52.8 & 37.9 & 39.8 \\
      + RL-Extra  & 36.9 & 18.5 & 59.9 & 35.5 & 41.2 & 54.2 & 35.6 & 40.3 \\
      + NExt (Ours)      & 35.3 & 20.3 & 60.8 & 35.8 & 40.2 & 54.2 & 36.5 & \underline{40.4} \\
      \midrule
      \multicolumn{9}{c}{\textit{Qwen2.5-7B-Instruct}} \\
      Backbone    & 46.6 & 31.0 & 70.3 & 36.3 & 53.6 & 61.3 & 52.2 & 50.2 \\
      + GRPO      & 49.4 & 31.1 & 70.7 & 41.0 & 59.4 & 61.6 & 54.4 & 52.5 \\
      + AlphaRL   & 47.4 & 31.8 & 70.0 & 39.5 & 58.7 & 61.8 & 54.4 & 51.9 \\
      + RL-Extra  & 51.1 & 31.3 & 70.6 & 41.0 & 58.7 & 64.2 & 53.9 & \underline{53.0} \\
      + NExt (Ours)      & 48.3 & 32.4 & 71.5 & 41.5 & 59.2 & 63.9 & 55.4 & \textbf{53.2} \\
      \midrule
      \multicolumn{9}{c}{\textit{Qwen2.5-14B-Instruct}} \\
      Backbone    & 64.2 & 32.3 & 83.1 & 44.2 & 64.7 & 72.4 & 64.9 & 60.8 \\
      + GRPO      & 63.6 & 33.3 & 83.8 & 47.7 & 69.3 & 71.5 & 68.8 & \textbf{62.6} \\
      + AlphaRL   & 62.8 & 32.1 & 85.2 & 45.1 & 67.5 & 73.6 & 66.9 & 61.9 \\
      + RL-Extra  & 61.1 & 35.2 & 83.4 & 49.1 & 69.8 & 71.0 & 65.4 & 62.1 \\
      + NExt (Ours)      & 65.1 & 32.1 & 84.7 & 47.9 & 69.8 & 71.1 & 67.0 & \underline{62.5} \\
      \bottomrule
      \end{tabular}
      \label{tab:mmlu_pro_2}
\end{table*}

\subsubsection{Performance on Other Domain Tasks}

To evaluate the adaptability of our method across different domains, we conduct experiments on MMLU-Pro~\cite{mmlu_pro} and GPQA Diamond~\cite{gpqa} using our approach.
These two tasks are multiple-choice benchmarks that cover a wide range of subjects, reflecting the model’s capabilities across different domains.
We present the experimental results on MMLU-Pro in Tables~\ref{tab:mmlu_pro_1} and~\ref{tab:mmlu_pro_2}, and report the model’s performance and computational cost on GPQA in Figure~\ref{fig:gpqa}.

From the experimental results, we observe that across various subjects, models trained with NExt achieve performance comparable to those trained with traditional RLVR methods. 
As we scale the model size from 1.5B to 14B, NExt consistently accelerates the training process, requiring only 250 RLVR steps to reach performance comparable to models trained with 400 RLVR steps.
Furthermore, Figure~\ref{fig:gpqa} compares the post-training performance and the training cost across different methods. On the GPQA task, NExt requires fewer GPU hours, and the computational resources used for extrapolation are significantly lower than those required for RLVR.
This indicates that the additional cost introduced by the extrapolation process in NExt is minimal and does not affect the overall training efficiency.
\section{Conclusion}

In this paper, we investigated the dynamics of the LLM rank-1 subspace during the RLVR process, observing that LoRA fine-tuning can better elicit the dominance of the rank-1 subspace and the rank-1 subspace evolves nonlinearly during the training process. 
Based on these critical insights, we proposed \textbf{NExt}, an approach that leverages the LoRA fine-tuning trajectories to perform extrapolation for the LLM rank-1 subspace.
Concretely, we first collected the intermediate checkpoints during LoRA-based RLVR training, and then computed three categories of LLM parameter differences.
Afterward, we utilized the computed deltas to construct the training dataset for the optimization trajectory predictor, and then leveraged the well-trained predictor to extrapolate the LLM parameters.
To evaluate the effectiveness and efficiency of our NExt, we conducted experiments to compare the competitive baselines, demonstrating that NExt can lead LLMs to achieve better performance within a small number of training steps and low computational overhead.
Furthermore, additional experiments show that NExt exhibits strong robustness and generalization ability.

In future work, we will further investigate the patterns of internal parameter updates during the RLVR process. These properties will help us better understand how the model evolves during training, thereby enabling parameter extrapolation to further reduce computational overhead and better facilitate test-time scaling.

\bibliography{iclr2026_conference}
\bibliographystyle{iclr2026_conference}

\end{document}